\documentclass{article}

\usepackage[utf8]{inputenc} 
\usepackage[T1]{fontenc}    
\usepackage{hyperref}       
\usepackage{url}            
\usepackage{booktabs}       
\usepackage{amsfonts}       
\usepackage{nicefrac}       
\usepackage{graphicx}

\newcommand{\hlf}{\textstyle \frac{1}{2} \displaystyle}
\newcommand{\reals}{{\rm I\kern-.17em R}}
\def\Reals{{\hbox{$\it I\hskip-3.6pt R$}}}

\def\Reals{{\hbox{$\it I\hskip-3.6pt R$}}}

\newcommand{\ie}{{\em i.e.}}

\newcommand{\etal}{{\em et al.}}
\newcommand{\Khat}{\widehat{K}}

\newcommand{\Xhat}{\widehat{X}}

\newcommand{\calL}{{\cal L}}

\newcommand{\ptilde}{\tilde{p}}

\newcommand{\keiko}{\stackrel{\triangle}{=}}

\title{From the Expectation Maximisation Algorithm to Autoencoded Variational Bayes}

\author{
  Graham W. Pulford\\
}

\date{October 2020}
\begin{document}
\maketitle

\begin{abstract}
Although the expectation maximisation (EM) algorithm was introduced in 1970, it remains somewhat inaccessible to machine learning practitioners due to its obscure notation, terse proofs and lack of concrete links to modern machine learning techniques like autoencoded variational Bayes. This has resulted in gaps in the AI literature concerning the meaning of such concepts like ``latent variables'' and ``variational lower bound,'' which are frequently used but often not clearly explained. The roots of these ideas lie in the EM algorithm. We first give a tutorial presentation of the EM algorithm for estimating the parameters of a $K$-component mixture density. The Gaussian mixture case is presented in detail using $K$-ary scalar hidden (or latent) variables rather than the more traditional binary valued $K$-dimenional vectors. This presentation is motivated by mixture modelling from the target tracking literature. In a similar style to Bishop's 2009 book, we present variational Bayesian inference as a generalised EM algorithm stemming from the variational (or evidential) lower bound, as well as the technique of mean field approximation (or product density transform). We continue the evolution from EM to variational autoencoders, developed by Kingma \& Welling in 2014. In so doing, we establish clear links between the EM algorithm and its variational counterparts, hence clarifying the meaning of ``latent variables.'' We provide a detailed coverage of the ``reparametrisation trick'' and focus on how the AEVB differs from conventional variational Bayesian inference. Throughout the tutorial, consistent notational conventions are used. This unifies the narrative and clarifies the concepts. Some numerical examples are given to further illustrate the algorithms. 
\end{abstract}


\tableofcontents
\newpage

\section{Introduction}
Finite mixture densities are a useful tool for modelling multi-dimensional, multi-modal data sets. For well separated modes, a finite mixture model that fits one component of a mixture density per mode can be viewed as an unsupervised learning or data clustering approach. Mixture density modelling is more general than data clustering since it fits a multi-modal probability density to the data, which can then be used for predictive purposes. Once the type of mixture model has been chosen (for instance, a Gaussian mixture model or GMM), which includes selecting the number of components, the fundamental problem is then to determine the parameter values of the mixture PDF such that the model ``fits'' the data. The maximum likelihood criterion is the yardstick for fitting in the context of mixture models, although a MAP (maximum {\em a posteriori}) criterion can also be used.

The monograph by Titterington \etal~\cite{Titterington} is one of the seminal works in this area, although it is not particularly accessible to AI practitioners. One of the motivations for this tutorial is to improve the accessibility of this theory to the AI community by simplifying the presentation of the EM algorithm in the next section. On a didactic level, we have avoided the usual practice of introducing binary hidden variables (taking values 0 or 1), and opted for discrete association variables $z_i$ where the event $z_i=j$ indicates that observation $x_i$ is attributed to mixture component $j$. This terminology is taken from the target tracking literature, where Gaussian mixtures have been employed since the mid-1970s, for instance, in the form of the probabilistic data association (PDA) algorithm \cite{Barshalom1}. Another motivating factor is to explain on a fundamental level what the role of hidden or latent variables is in the context of the parameter estimation problem for mixture densities.

The final goal of this tutorial is to show the evolution of EM, through its variational generalisations including the mean-field theory approach to variational Bayesian optimisation (as described in \cite{Bishop2009}). We end with a treatment of the well-cited autoencoded variational Bayes (AEVB) algorithm of Kingma and Welling \cite{Kingma2014}, which appeared after the 2009 edition of Bishop's book. All of these techniques make use in one form or another of a lower bound on the data likelihood, either via Baum's auxiliary function, or by a functional of the latent variable prior distribution. Despite the uptake of the popular AEVB technique, certain details of the algorithm deserve more attention. We provide a discussion of the so-called ``reparametrisation trick'' and focus on how the AEVB differs from conventional variational inference based on the variational lower bound (VLB). In closing, we formalise the application of the reparametrisation trick in the AEVB algorithm.

\section{Expectation Maximisation}
\subsection{Parameter Estimation for Gaussian Mixture Densities}\label{EMalg}
Parameter estimation for finite mixture models is inherently ill conditioned: methods that seek directly to maximise the likelihood function exhibit poor convergence. Moreover, the maximum likelihood objective function is intractable since the number of terms is ${\rm O}(K^N)$ where $K$ is the number of mixture components and $N$ is the number of data samples. This is clearly exponential in the number of samples. The Expectation Maximisation (EM) algorithm is the main method for obtaining mixture parameter estimates in a numerically tractable way. In essence, it introduces a set of auxiliary variables called ``hidden'' or ``latent'' variables, via which the joint PDF is most easily expressed. Since these variables are unknown, an iterative approach is adopted wherein the hidden variables are ``averaged out'' of the joint density at each iteration. The parameters are re-estimated by optimising the resulting averaged density and the process is repeated until convergence.

The following treatment of the EM algorithm adopts notation from chapter 9 of \cite{Bishop2009}, which is more in line with the notation typically used in the AI literature than the notation found in the statistics literature.
\begin{itemize}
    \item Incomplete data $X=\{x_1,\ldots,x_N\},~x_i\in{R}^n$, also called unlabelled or uncategorised data, depending on the problem context.
    \item Missing data $Z=\{z_1,\ldots,z_N\}$, also called hidden or latent variables.
    \item Complete data $Y=(X,Z)$, the union of the incomplete and missing data.
    \item Mixture parameters $\Theta$. In the case of a $K$-component multivariate Gaussian mixture - the set of parameters of the mixture consisting of (scalar) weights $\pi_i$, mean vectors $\mu_i$,  covariance matrices $P_i$.
\end{itemize}
A basic assumption of the EM algorithm is that each data point $x_i$ is from one and only one component $j$ of a finite mixture distribution. The missing information refers to the component of the mixture pertaining to each observation. In the case of a finite mixture distribution, each element of the missing data vector is an indicator (or label) of the associated mixture component, one per data point. In the case of a Gaussian mixture, the PDF takes the form:
\[
{\rm p}(x)=\sum_{k=1}^K \pi_k {\rm N}(x; \mu_k,P_k)
\]
where ${\rm N}(x; \mu,P)$ is a multivariate Gaussian PDF in the variable $x$, with mean vector $\mu$ and covariance matrix $P$, \ie,
\begin{equation}\label{gaussian}
{\rm N}(x; \mu,P)=(2\pi)^{-n_x/2}({\rm det}P)^{-\hlf}\exp\{-\hlf(x-\mu)^TP^{-1}(x-\mu)\}
\end{equation}
where $n_x$ is the dimension of $x$.
The mixture weights satisfy $\pi_k\geq 0$ $\forall k$ and
\[
\sum_{k=1}^K \pi_k=1,
\]
from which it follows that $\pi_k={\rm Pr}(x\leftrightarrow\mbox{component k})$. 
The set of parameters of the Gaussian mixture is denoted $\Theta=\{\pi_k, \mu_k,P_k\}_{k=1}^K$. As previously mentioned, the event $z_i=j$, $j\in\{1,\ldots,K\}$ signifies that observation $x_i$ is attributed to mixture component $j$, or, in other words, the conditional distribution of $x_i$ given $z_j$ is
\[
x_i|z_i\sim {\rm N}(x;\mu_j,P_j)
\]
The standard criterion for estimating the parameters $\{\Theta\}$ of a mixture distribution from incomplete observations $X$ using the EM algorithm is maximum likedlihood (ML), although other criteria, such as maximum a posteriori can also be used. In the case of ML, the parameters are obtained by optimising the objective function:
\begin{equation}\label{em1}
    \hat{\Theta}=\arg\max_{\Theta} {\rm p}(X|\Theta)
\end{equation}
Even in the case of independent observations (white noise), the likelihood function for a mixture with unlabelled or uncategorised observations (\ie, when the labels $z_i$ are unavailable) takes the form
\begin{equation}\label{pxtheta}
{\rm p}(X|\Theta)=\prod_{i=1}^N {\rm p}(x_i|\Theta)=\prod_{i=1}^N
\sum_{k=1}^K \pi_k {\rm N}(x_i; \mu_k,P_k)
\end{equation}
Taking the log of the likelihood affords only  limited simplification, {\em viz.}
\begin{equation}\label{logpxtheta}
\log{\rm p}(X|\Theta)=\sum_{i=1}^N
\log\left\{ \sum_{k=1}^K \pi_k {\rm N}(x_i; \mu_k,P_k)\right\}
\end{equation}
No further simplification of the log likelihood is possible in general, and we are left with a numerically ill-conditioned multi-dimensional optimisation problem to obtain the mixture parameters. The ill conditioning arises from the possibility that one or more of the mixture means can be exactly equal to one of the data points $x_i$. In this case the conditional Gaussian PDF behaves as ${\rm det}(P_j)^{-\hlf}$, which tends to infinity when the determinant of the covariance becomes small. Illustrations of this phenomenon appear in \cite{Bishop2009}.

The preceding arguments motivate the expectation-maximisation (EM) algorithm, first introduced by Baum \etal~\cite{Baum2,Dempster}. Instead of the direct ML optimisation in (\ref{em1}), Baum et al. introduced the {\em auxiliary function} $Q(\Theta,\Theta_0)$, which is defined as
\begin{eqnarray}\label{baum}
    Q(\Theta,\Theta_0)&=&{\rm E}_Z[\log {\rm p}(X,Z|\Theta) | X,\Theta_0]\\
    &=&\int \log({\rm p}(X,Z|\Theta))\,{\rm p}(Z | X,\Theta_0)\,dZ\nonumber
\end{eqnarray}
where the expectation is with respect to the joint density of the hidden variables $Z$ given the incomplete data $X$ and some nominal parameter estimates $\Theta_0$. Since $\varphi(\cdot)=\log(\cdot)$ is a convex function, one can apply Jensen's inequality $\varphi({\rm E}[U])\leq{\rm E}[\varphi(U)]$ for any random variable $U$, to show that
\[
\Theta_1=\arg\max_{\Theta} Q(\Theta,\Theta_0) \Rightarrow {\rm p}(X|\Theta_1) \geq {\rm p}(X|\Theta_0)
\]
This makes it feasible to replace the maximisation (\ref{em1}) by an iterative procedure that seeks at each stage (or ``pass'') $p$ the maximum of Baum's auxiliary function. The resulting EM algorithm consists of a loop with two steps as follows
\begin{enumerate}
    \item Expectation (E-step) :
    \[
Q(\Theta,\Theta_{p-1})={\rm E}_Z[\log {\rm p}(X,Z|\Theta) | X,\Theta_{p-1}]
    \]
    \item Maximisation (M-step) :
 \[
\Theta_p=\arg\max_{\Theta} Q(\Theta,\Theta_{p-1})
 \]
\end{enumerate}
Note that the algorithm requires multiple passes through the entire data set $X$ to continually refine the parameter estimates $\Theta_p$. There is no guarantee that the algorithm will converge to the ML estimator since there can be multiple stationary points in the likelihood function, and, depending on the initial parameter estimates, the EM algorithm may converge to any of these.

The utility of the EM algorithm rests on the suppositions that (i) the E-step is explicit and (ii) the M-step is easier to calculate than direct maximisation of the likelihood function. For mixture distributions from the exponential family, which includes finite Gaussian mixtures, this is the case. In such cases, the conditioning on the hidden variables allows the joint data likelihood (for independent observations) to be expressed as 
\[
{\rm p}(Y|\Theta)=\prod_{i=1}^N{\rm p}(x_i,z_i|\Theta)=\prod_{i=1}^N{\rm p}(x_i|z_i,\Theta)
\Pr(z_i|\Theta)=\prod_{i=1}^N \pi_{z_i}{\rm N}(x_i;\mu_{z_i},P_{z_i})
\]
This at once makes clear the rationale for forming the complete data likelihood, namely, that the latter quantity is expressible as a product of Gaussian PDFs and mixture weights. Taking the logarithm of this results in a manageable expression
\[
\log\,{\rm p}(Y|\Theta)=\sum_{i=1}^{N}\log\,\pi_{z_i}+\sum_{i=1}^{N}\log\,{\rm N}(x_i;\mu_{z_i},P_{z_i})
\]
 whose expectation with respect to the hidden variables can be explicitly obtained (for the E-step). Contrast this with the incomplete data likelihood in (\ref{pxtheta}), the number of terms in which increases exponentially with the length of the data sequence.

In order to derive the EM algorithm, we must now calculate Baum's auxiliary function and then maximise it with respect to the parameters. We present a detailed derivation of the E-step below and then specialise it to the Gaussian mixture case.

Starting from (\ref{baum}), and noting that the hidden variables are discrete, we have
\begin{eqnarray}
    Q(\Theta,\Theta_0)&=&{\rm E}_Z[\log {\rm p}(X,Z|\Theta) | X,\Theta_0]\label{baum2a}\\
    &=&\sum_{z_1=1}^{K}\cdots\sum_{z_N=1}^{K}\log{\rm p}(X,Z|\Theta)\Pr(Z|X,\Theta_0)\label{baum2b}
\end{eqnarray}
Factorisation of the joint density of the complete data is possible due to the assumed independence of the data samples. The right hand side of (\ref{baum2b}) therefore decomposes into a sum of log-PDF terms each of which involves a product of probabilities of the hidden variables. To simplify the resulting calculations, we define
\begin{eqnarray}
g(z_i,x_j) &=& \log{\rm p}(z_i,x_j|\Theta)\\
h(z_i|x_j) &=& \Pr(z_i|x_j,\Theta_0)\nonumber
\end{eqnarray}
With this shorthand, Baum's auxiliary function becomes
\begin{equation}\label{baum3}
Q(\Theta,\Theta_0)=\sum_{z_1=1}^{K}\cdots\sum_{z_N=1}^{K}\sum_{i=1}^{N}g(z_i,x_i)
\prod_{j=1}^N h(z_j|x_j)
\end{equation}
Upon expanding the inner sum over $i=1:N$ and rearranging we have
\begin{eqnarray*}
Q(\Theta,\Theta_0) &=& \sum_{z_1=1}^{K}g(z_1,x_1)h(z_1|x_1)\sum_{z_2=1}^{K}\cdots\sum_{z_N=1}^{K}\prod_{j\neq 1} h(z_j|x_j)+\cdots\\
 &&+~\sum_{z_N=1}^{K}g(z_N,x_N)h(z_N|x_N)\sum_{z_1=1}^{K}\cdots\sum_{z_{N-1}=1}^{K}\prod_{j\neq N} h(z_j|x_j)
\end{eqnarray*}
which consists of $N$ $N$-fold sum-products. Now consider the $(N-1)$-fold sum in the last part of the above expression:
\[
\sum_{z_1=1}^{K}\cdots\sum_{z_{N-1}=1}^{K}\prod_{j=1}^{N-1} h(z_j|x_j)=
\sum_{z_1=1}^{K}\cdots\sum_{z_{N-2}=1}^{K}\prod_{j=1}^{N-2}h(z_j|x_j)
\sum_{z_{N-1}=1}^{K}h(z_{N-1}|x_{N-1})
\]
The definition of the hidden variables means that for any one of the $z_i$
\[
\sum_{z_i=1}^{K}h(z_i|x_i)=\sum_{k=1}^{K}h(z=k|x_i)=\sum_{k=1}^{K}\Pr(z=k|x_i,\Theta_0)=1
\]
So it eventuates that
\[
\sum_{z_1=1}^{K}\cdots\sum_{z_{N-1}=1}^{K}\prod_{j=1}^{N-1} h(z_j|x_j)=
\sum_{z_1=1}^{K}\cdots\sum_{z_{N-2}=1}^{K}\prod_{j=1}^{N-2}h(z_j|x_j)
\]
and, by reverse induction on $z_n$ for $n=N-2,N-3,\ldots,2,1$ it follows (somewhat miraculously) that
\[
\sum_{z_1=1}^{K}\cdots\sum_{z_{N-1}=1}^{K}\prod_{j=1}^{N-1} h(z_j|x_j)=
\sum_{z_1=1}^{K}h(z_1|x_1)=1
\]
Returning to (\ref{baum3}), we now have
\begin{eqnarray}
Q(\Theta,\Theta_0) &=& \sum_{z_1=1}^{K}g(z_1,x_1)h(z_1|x_1)+\cdots+\sum_{z_N=1}^{K}g(z_N,x_N)h(z_N|x_N)\nonumber\\
 &=& \sum_{n=1}^{N}\sum_{k=1}^{K}g(z=k,x_n)h(z=k|x_n)\nonumber\\
 &=& \sum_{n=1}^{N}\sum_{k=1}^{K}\Pr(z=k|x_n,\Theta_0)\log{\rm p}(z=k,x_n|\Theta)\label{baum4}
\end{eqnarray}
Examining the second term in the double sum, we notice that
\begin{eqnarray*}
\log{\rm p}(z=k,x_n|\Theta) &=& \log\left(\Pr(z=k|\Theta){\rm p}(x_n|z=k,\Theta)\right)\\
 &=& \log \pi_k+\log {\rm p}(x_n|z=k,\Theta)
\end{eqnarray*}
Applying Bayes' rules to the first term in (\ref{baum4}) we can show that
\[
\Pr(z=k|x_n,\Theta_0)=\frac{\pi^{(0)}_k{\rm p}(x_n|\theta^{(0)}_k)}{
\sum_{j=1}^{K}\pi^{(0)}_j{\rm p}(x_n|\theta^{(0)}_j)}
\]
where we have defined $\theta^{(p)}_k=\{\pi^{(p)}_k, \mu^{(p)}_k,P^{(p)}_k\}$ as the set of parameters for mixture component $k$ at pass $p$ and ${\rm p}(x|\theta_j)$ is shorthand for the PDF of component $j$ of the mixture density and the superscript denotes quantities computed at the relevant pass. From (\ref{baum4}), the E-step for the EM algorithm is therefore given for a general finite mixture as
\begin{equation}\label{estep}
Q(\Theta,\Theta_{p-1})=\sum_{n=1}^{N}\sum_{k=1}^{K}w_{nk}(\Theta_{p-1})\left(\log\,\pi_k+\log\,{\rm p}(x_n|\theta_k)
\right)
\end{equation}
where the ``weights'' defined for $n=1,\ldots,N$ and $k=1,\ldots,K$ are given by
\[
w_{nk}(\Theta_{p-1})=
\frac{\pi^{(p-1)}_k{\rm p}(x_n|\theta^{(p-1)}_k)}{
\sum_{j=1}^{K}\pi^{(p-1)}_j{\rm p}(x_n|\theta^{(p-1)}_j)}
\]
In the Gaussian mixture case, the weights are obtained as
\[
w_{nk}(\Theta_{p-1})=\frac{\pi^{(p-1)}_k {\rm N}(x_n;\mu^{(p-1)}_k,P^{(p-1)}_k)}{\sum_{j=1}^{K}\pi^{(p-1)}_j{\rm N}(x_n;\mu^{(p-1)}_j,P^{(p-1)}_j)}
\]
where the PDF was defined in (\ref{gaussian}). In this case, the E-step takes the form
\begin{eqnarray*}
Q(\Theta,\Theta_{p-1}) &=& \sum_{n=1}^{N}\sum_{k=1}^{K}w_{nk}(\Theta_{p-1})\left(\log\,\pi_k-\frac{n_x}{2}\log(2\pi)-\hlf\log({\rm det}P_k)\right.\\
 &&~ \left.-\hlf(x_n-\mu_k)^TP_k^{-1}(x_n-\mu_k)\right)
\end{eqnarray*}

The updated mixture weights, means and covariance matrices are obtained from the M-step, subject to the constraint that the mixture weights sum to unity. The update for the mixture weights is easy to derive and does not depend on the type of mixture PDF. The updates for the mixture means and covariances require a bit more work and we refer the reader to \cite{Bilmes} for the detailed derivations. The result is:
\begin{eqnarray*}
\pi^{(p)}_k &=& \frac{1}{N}\sum_{n=1}^{N}w_{nk}(\Theta_{p-1}) \\
\mu^{(p)}_k &=& \frac{\sum_{n=1}^{N}w_{nk}(\Theta_{p-1})x_n}{\sum_{n=1}^{N}w_{nk}(\Theta_{p-1})}\\
P^{(p)}_k &=& \frac{\sum_{n=1}^{N}w_{nk}(\Theta_{p-1})(x_n-\mu^{(p)}_k)(x_n-\mu^{(p)}_k)^T}{\sum_{n=1}^{N}w_{nk}(\Theta_{p-1})}
\end{eqnarray*}
The loop of the EM algorithm is completed by setting $\Theta_p=\{\pi^{(p)}_k, \mu^{(p)}_k,P^{(p)}_k\}_{k=1}^K$ and $p\leftarrow p+1$ and iterating until convergence. Note that the number of mixture components $K$ must be chosen beforehand.

\subsection{2-D Numerical Example of the EM Algorithm}

\begin{figure}\label{fig1}
\center
   \includegraphics[height=6cm]{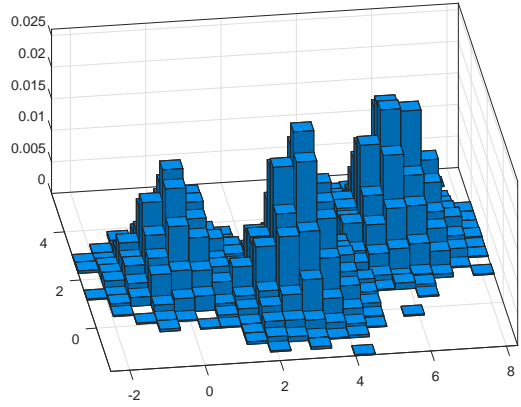}
\caption{2-D Histogram of 3-component Gaussian mixture data input to EM algorithm.}
\end{figure}

A numerical example of the expectation maximisation algorithm is presented in this section. The truth data $X$ are generated from the following Gaussian mixture model (see Fig. \ref{fig1}), noting that any model (including real data) can potentially be used, but by using a GMM it is possible to check the accuracy of the EM estimates.

\subsection*{Data Generation Parameters}
\begin{enumerate}
\item Number of components: $K=3$.
\item Number of discrete samples: $N=5000$. IID samples.
\item Mixture weights: $\pi_1=0.25$, $\pi_2=0.40$, $\pi_3=0.35$.
\item Mixture means $\mu_1=[0,~2]^T$, $\mu_1=[3,~1]^T$, $\mu_1=[6,~3]^T$.
\item Mixture covariances $P_1=P_2=P_3={\rm diag}\,[0.5,~0.5]$.
\item Stopping criterion: whichever occurs sooner of (i) number of passes $p$ reaches 50 or (ii) the following convergence condition is satisfied:
\[
\left|\,\sum_{n=1}^{N}\log{\rm p}(X_n|\Theta_p)-\sum_{n=1}^{N}\log{\rm p}(X_n|\Theta_{p-1})\right|<10^{-3}
\]
\end{enumerate}

\subsection*{Implementation Parameters}
The EM algorithm described at the end of section \ref{EMalg} is implemented with initialisation carried out according to the recipe below.
\begin{enumerate}
\item Input: 2-D data sequence $X$, assumed number of components $\Khat$, initial mixture parameter estimate $\Theta_0$, (maximum) number of passes $N_P=50$.
\item The empirical data region is obtained from $X$ as $[x_{min},x_{max}]\times[y_{min},y_{max}]$, where $x$ and $y$ are the 1st and 2nd components of $X$. No outlier detection is applied.
\item Initial mixture weights $\pi_k^{(0)}=1/\Khat$, $k=1,\ldots,\Khat$.
\item Define $r(\Khat)={\rm ceil}(\sqrt{\Khat})$ where ``ceil'' is the ceiling function.
\item Divide the data region into $r(\Khat)\times r(\Khat)$ equal-sized cells.
\item Randomly choose $\Khat$ cells $C_1,\ldots,C_{\Khat}$
\item Set initial mixture mean estimates $\mu_k^{(0)}=\mbox{centre of cell }C_k$, $k=1,\ldots,\Khat$.
\item Define $\sigma_x=(x_{max}-x_{min})/6$ and $\sigma_y=(y_{max}-y_{min})/6$
\item Set initial mixture covariance estimates $P_k^{(0)}={\rm diag}\,[\sigma_x^2,\sigma_y^2]$, $k=1,\ldots,\Khat$.
\end{enumerate}

\subsection*{Numerical Results}
In order to demonstrate the effect of the assumed number of mixture components, a range of $\Khat$ values were chosen from 2 to 6, bearing in mind that the true value is $K=3$. The random initialisation carried out as described above. We give the results in both tabular and graphical form. For tabular results, the mixture weights and means are given but the covariance matrices are omitted (except for $\Khat=3$). For graphical results, the raw $X$ samples are plotted as a point cloud (with multiple colours). For each mixture component at each pass of the EM algorithm, we plot the 1-sigma ellipse corresponding to the mean and covariance of the given mixture component. Each mixture component is colour coded according to the following scheme: component 1 black; component 2 blue; component 3 red; component 4 magenta; component 5 cyan.

\begin{figure}
\center
   \includegraphics[height=8.5cm]{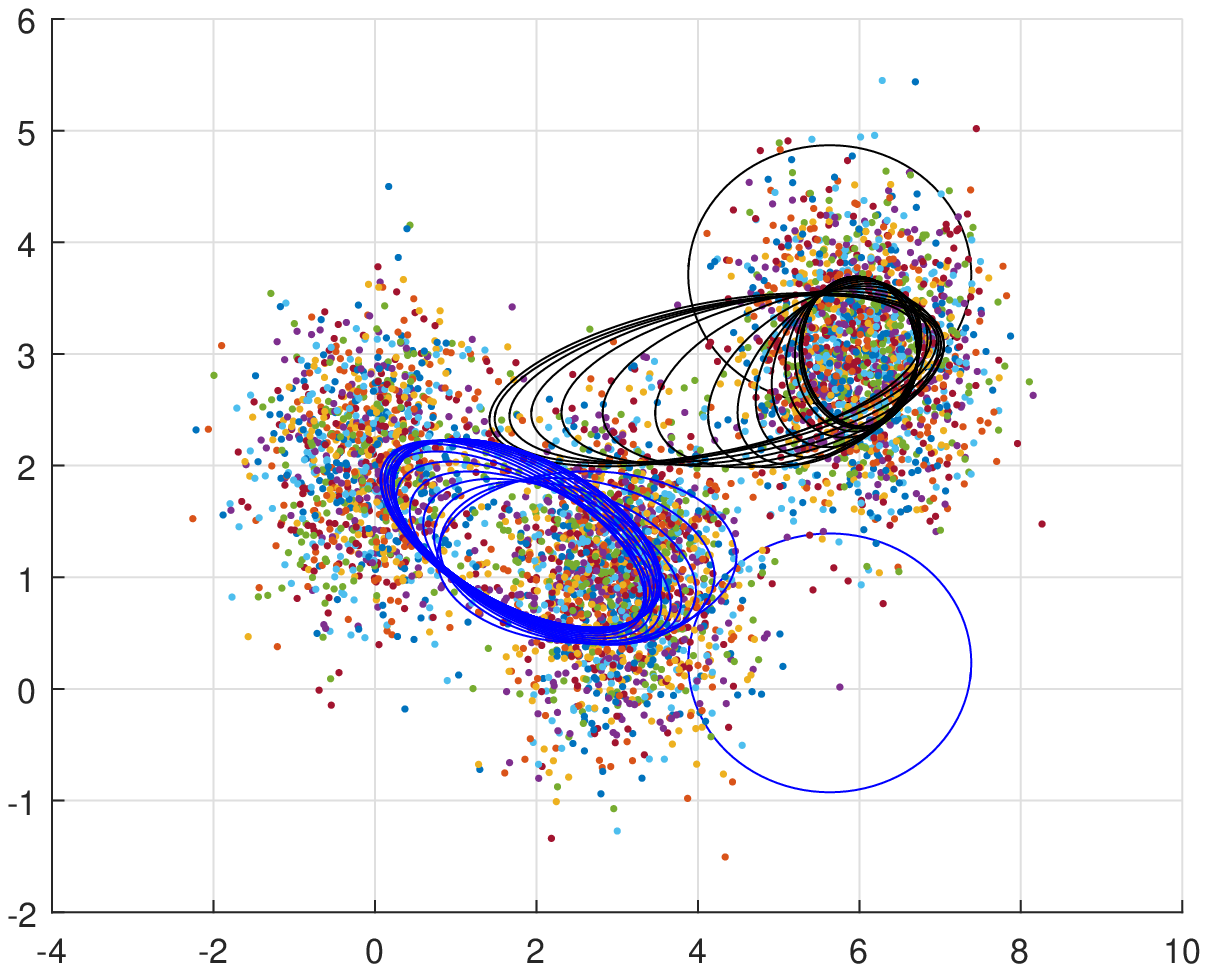}
\caption{EM algorithm results for $\Khat=2$ components. Raw data shown as point cloud. Coloured ellipses correspond to evolution of EM mixture component estimates (larger at start). Colour ordering of mixture components: black, blue. Weights and means given in Table \ref{Tab1}.}
\label{fig2}
\end{figure}

Tables and graphs have been relegated to the end of the document.  We have not calculated explicit performance figures, which at any rate require Monte Carlo testing over different data realisations, number of samples and parameter settings. Instead, a qualitative summary is given. For $\Khat=2$, shown in Fig. \ref{fig2}, the EM algorithm estimate for true mixture component 3 converged satisfactorily after 20 passes but the other mixture estimate wandered into a region between true mixture components 1 and 2, which are the most closely spaced modes of the trimodal truth data.

\begin{figure}
\center
   \includegraphics[height=9cm]{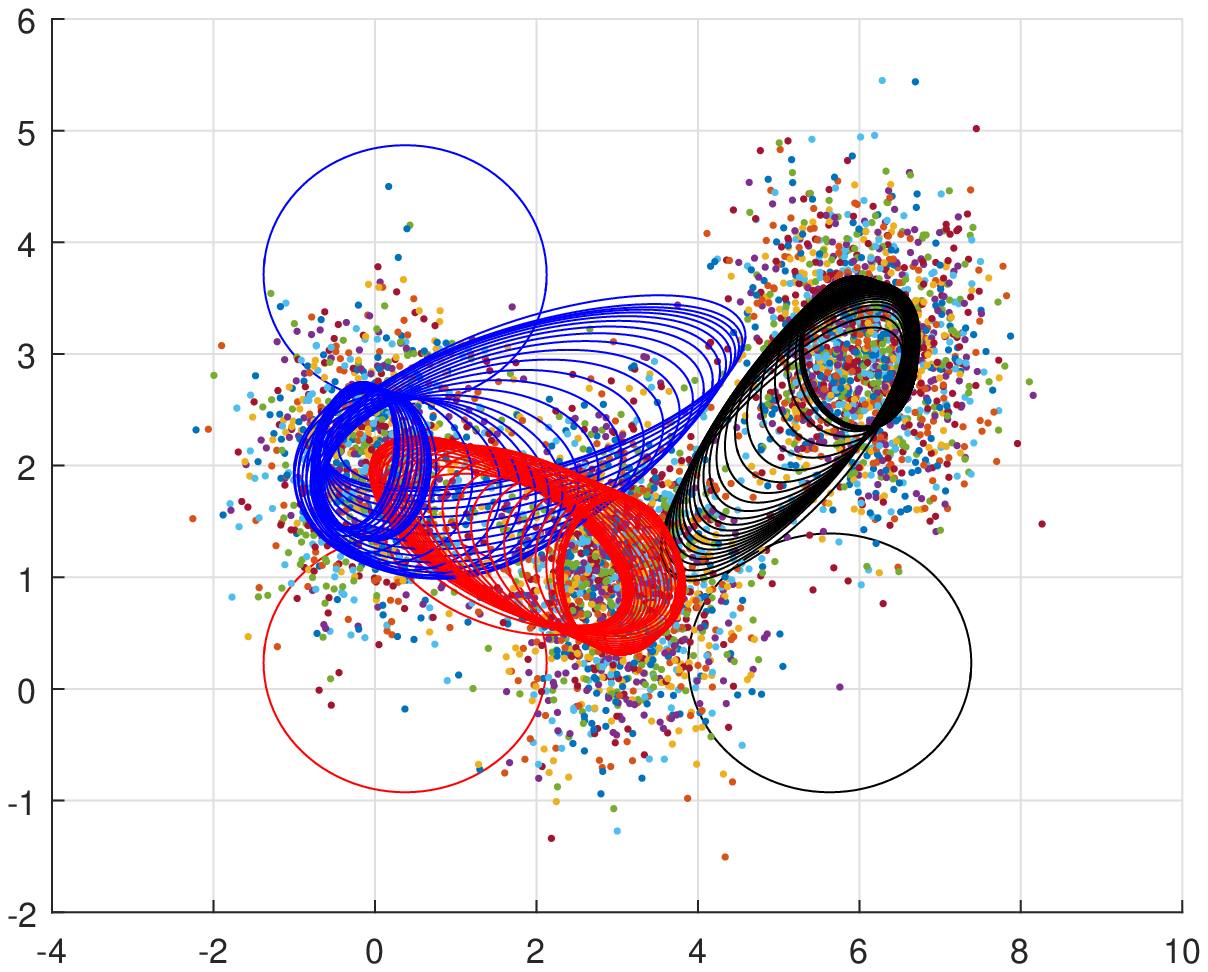}
\caption{EM algorithm results for $\Khat=3$ components. Stop at 46 passes. Raw data shown as point cloud. Coloured ellipses correspond to evolution of EM mixture component estimates (larger at start). Colour ordering of mixture components: black, blue, red. Weights and means given in Table \ref{Tab2}.}
\label{fig3}
\end{figure}

For $\Khat=3$, shown in Fig. \ref{fig3}, convergence to tolerance 0.001 in absolute log likelihood error was observed after 46 passes. At this stage, the 3 mixture component estimates were within 1\% relative error in all components of all mixture means. The mixture weights were accurate to less than 3\% relative error. The covariances (which are provided in this case) also closely match the true values.

\begin{figure}
\center
   \includegraphics[height=9cm]{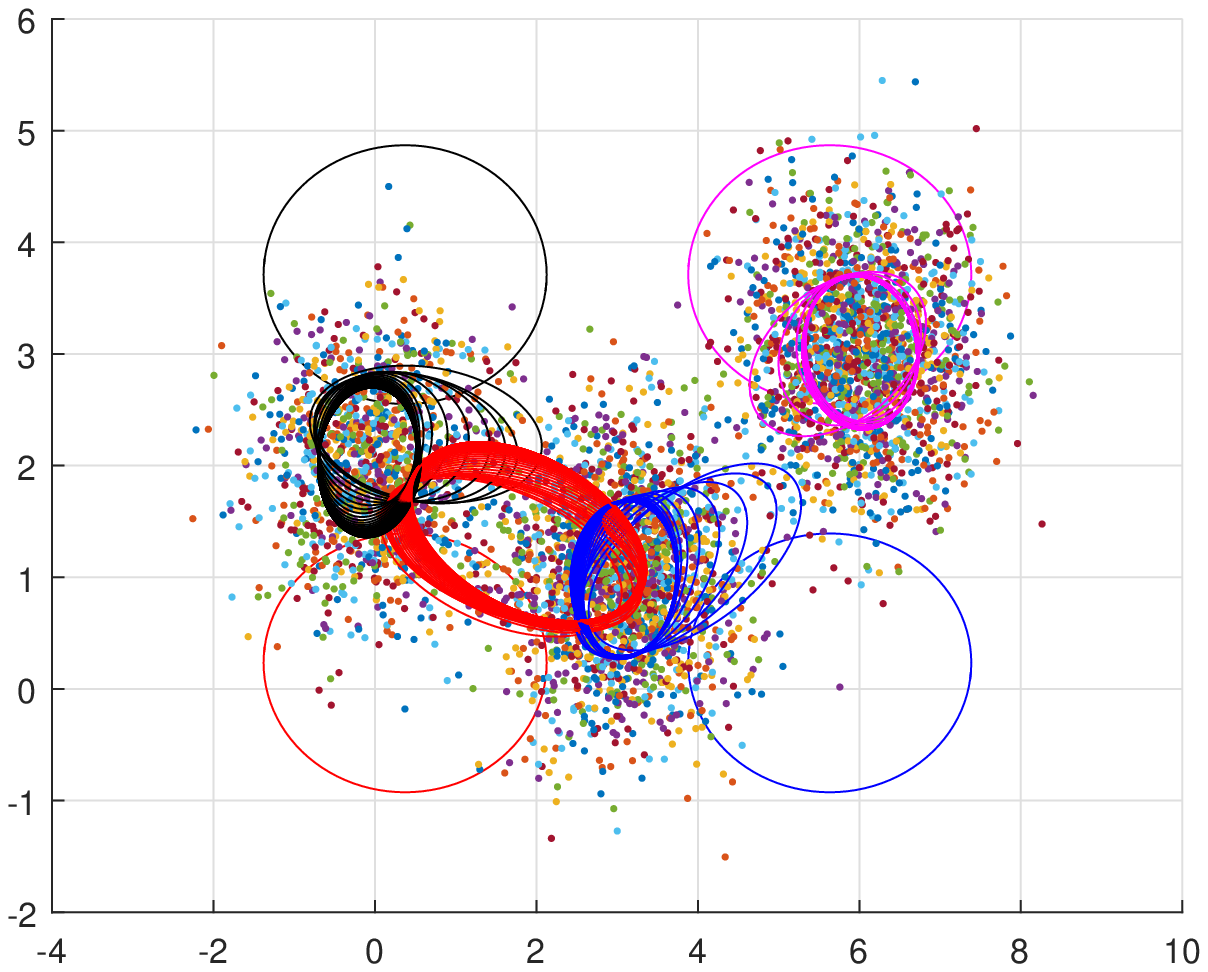}
\caption{EM algorithm results for $\Khat=4$ components. Stop at 50 passes. Raw data shown as point cloud. Coloured ellipses correspond to evolution of EM mixture component estimates (larger at start). Colour ordering of mixture components: black, blue, red, magenta. Weights and means given in Table \ref{Tab3}.}
\label{fig4}
\end{figure}

For $\Khat=4$, the iteration was stopped at pass 50 before achieving convergence to the preset tolerance. At this point, three of the mixture components (numbers 1, 2 and 4) had satisfactorily converged (at least in mean, although not yet in covariance. The third mixture component (shown in red in Fig. \ref{fig4}) still has a large covariance ellipse and is slowly settling the region between true mixture components 1 and 2. This component also has a relatively small mixture weight of 0.078.

\begin{figure}
\center
   \includegraphics[height=9cm]{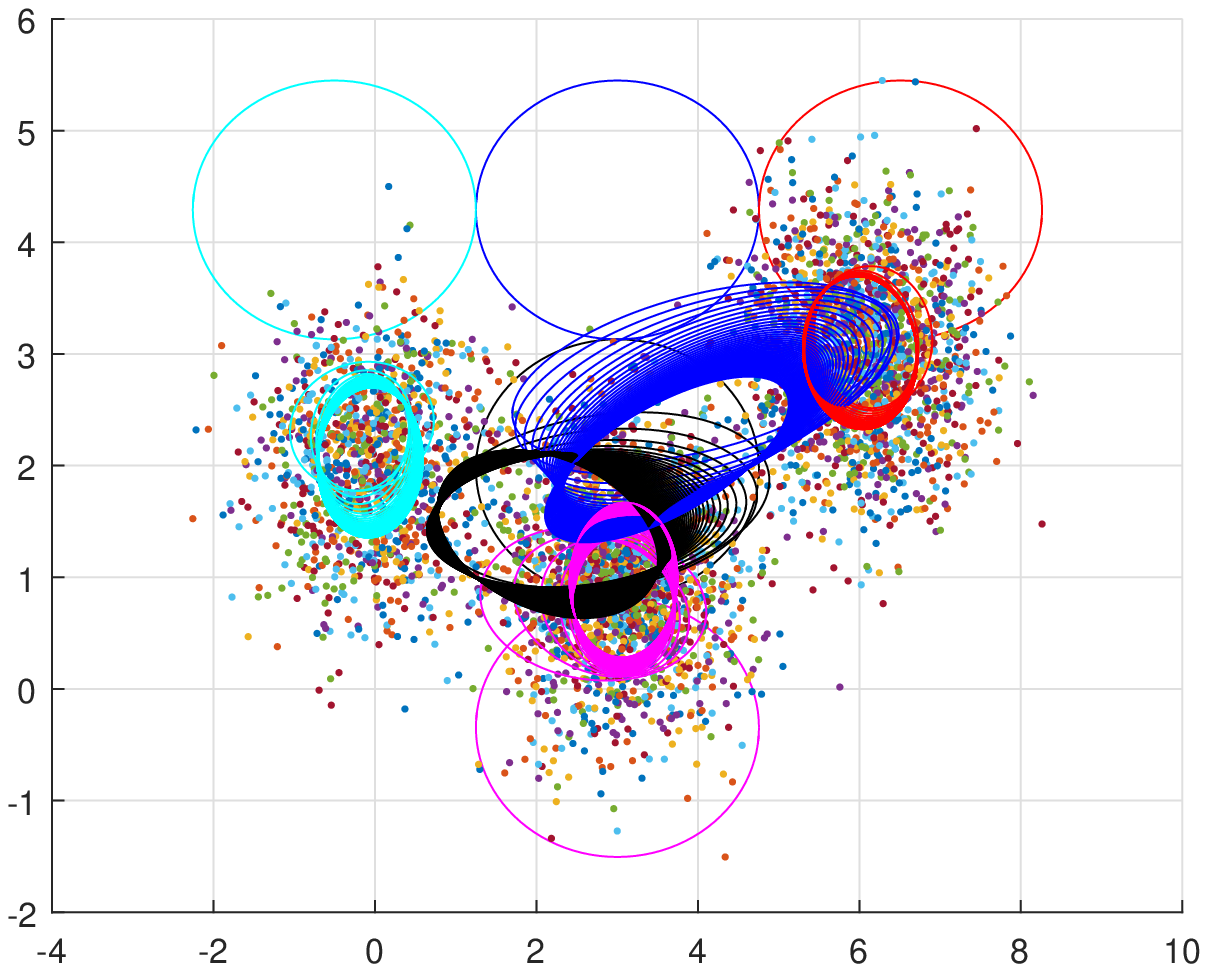}
\caption{EM algorithm results for $\Khat=5$ components. Raw data shown as point cloud. Coloured ellipses correspond to evolution of EM mixture component estimates (larger at start). Colour ordering of mixture components: black, blue, red, magenta, cyan. Weights and means given in Table \ref{Tab4}.}
\label{fig5}
\end{figure}

For $\Khat=5$, the iteration was also stopped at pass 50 before convergence. At this point, three of the mixture components (numbers 3, 4 and 5) had satisfactorily converged (at least in mean, although not yet in covariance. The first mixture component (shown in black in Fig. \ref{fig5}) is slowly settling the region between true mixture components 1 and 2, while the second mixture component (shown in blue in Fig. \ref{fig5}) is slowly settling the region between true mixture components 2 and 3.

For $\Khat=6$, whose results are not included, excess mixture components were observed to converge to regions either between true mixture modes or inside the same true mixture mode.

\subsection{Generalised EM Algorithm}\label{gemalg}
A more general version of the EM algorithm can be developed as explained in \cite{Neal99} and in chapter 9.4 of \cite{Bishop2009} that makes use of the so-called variational lower bound (VLB) and calls for a functional or variational maximisation in the E-step of the EM algorithm. This form is a pre-requisite for understanding variational Bayes inference in section \ref{VB} and the autoencoded variational Bayes algorithm described in section \ref{aevb}.

Notation (following \cite{Bishop2009}, section 9.4):

\begin{itemize}
    \item Vector of parameters $\Theta$ to be estimated.
    \item $X=\{x_1,\ldots,x_N\},~x_i\in\Reals^n$ incomplete data, \ie, the available observed or measured data.
    \item $Z$ set of latent, hidden or missing variables.
    \item $q(Z)$ unknown prior joint PDF of the latent variables.
    \item $\calL[q,\Theta]$ functional of the latent variable PDF and the parameters. This quantity is sometimes referred to as $F(q,\Theta)$.
    \item ${\rm KL}(q||p)$  Kullback-Liebler divergence: a similarity measure between two PDFs $q$ et $p$ given by:
    \[
    {\rm KL}(q||p)=-\int\log\left(\frac{{\rm p}(Z|X,\Theta)}{q(Z)}\right)q(Z)\,dZ
    \]
    The KL divergence is always non-negative. It is zero if and only if $q\equiv p$, that is, $p(Z)=q(Z)$ except on a set of measure zero.
\end{itemize}
For any probability density function (PDF) $q$, we can write:
\begin{equation}\label{logpkl}
    \log {\rm p}(X|\Theta)=\calL[q,\Theta]+{\rm KL}(q||p)
\end{equation}
where the {\em latent variable functional} is defined, in the case of discrete random variables, by:
\begin{equation}
    \calL[q,\Theta]={\rm E}_Z\log\left(\frac{{\rm p}(X,Z|\Theta)}{q(Z)}\right)=\sum_{Z}\log\left(\frac{{\rm p}(X,Z|\Theta)}{q(Z)}\right)q(Z)
\end{equation}
and, in the case of continuous random variables, by:
\begin{equation}\label{calL}
    \calL[q,\Theta]=\int\log\left(\frac{{\rm p}(X,Z|\Theta)}{q(Z)}\right)q(Z)\,dZ
\end{equation}
Note: neither the discrete sum nor the integral are easy to evaluate in general. For instance, the sum may involve a number of terms that grows exponentially with the number of latent variables. A simplification results when $Z$ assigns a single component of the mixture density to each data point in $X$.

Since the K-L divergence is non-negative, it follows from (\ref{logpkl}) that:
\[
\calL[q,\Theta]\leq \log {\rm p}(X|\Theta)
\]
with equality if and only if $q(Z)\equiv{\rm p}(Z|X,\Theta)$ (except on a set of $Z$ of measure zero). $\calL[q,\Theta]$ is therefore a {\em lower bound} on the log-likelihood function $\log {\rm p}(X|\Theta)$.

It is also clear from the definition of the latent variable functional in (\ref{calL}) that:
\begin{eqnarray}
    \calL[q,\Theta]&=&{\rm E}_Z{\rm p}(X,Z|\Theta)-{\rm E}_Z\log q(Z)\\
    &=&Q(\Theta,\Theta_0)-{\rm E}_Z\log q(Z)
\end{eqnarray}
where the last line assumes $Z\sim q(Z;\Theta_0)$. This establishes a link between the functional $\calL[q,\Theta]$ and Baum's auxiliary function $Q(\cdot,\cdot)$ (\ref{baum}).

In the same way as for the conventional EM algorithm, we can attempt to maximise the log-likelihood of the data indirectly via the the latent variable functional $\calL[q,\Theta]$.
\begin{enumerate}
    \item  Variational E-step :
    \[
    q_k=\arg\max_{q}\calL[q,\Theta_{k-1}]
    \]
    \item M-step :
 \[
\Theta_k=\arg\max_{\Theta} \calL[q_k,\Theta]
 \]
\end{enumerate}
The first stage of this {\em generalised} EM algorithm involves a maximisation of the latent variable functional $\calL[q,\Theta]$ over a space of probability density functions $q(\cdot)$, which is a variational optimisation. This type of problem can usually only be solved by assuming a parametric form for the PDF $q(\cdot)$, or by representing the PDF by a set of randomly drawn samples, as done in particle filtering (PF) approaches \cite{Gordon93} or Markov chain Monte Carlo (MCMC) \cite{Metropolis}.

\section{Variational Bayes}\label{VB}
\subsection{Variational Lower Bound}
As noted by Bishop \cite{Bishop2009}, variational methods originated in the 18th century in the context of classical mechanics and the well known works of Euler and Lagrange. The formulation assumes a functional, that is a mapping from a space of continuously-differentiable (smooth) functions to the real numbers, for which the function of the independent variables achieving the optimum is sought. The problem usually includes initial and terminal conditions (or boundary conditions in the multi-dimensional case), and the functional can additionally depend on derivatives up to a given order. For example, consider a functional operating on $f(x)$ and its first derivative:
\[
J[f]=\int_{x_1}^{x_2} L(x,f(x),f'(x))\,dx
\]
Calculus of variations gives the solution for optimising $J[f]$ in terms of the Euler-Lagrange equation or functional derivative:
\[
\frac{\delta J}{\delta f(x)}=\frac{\partial L}{\partial f}-\frac{d}{dx}\left(\frac{\partial L}{\partial f'}\right)=0
\]
Note that, apart from initial and terminal conditions such as $f(x_1)=f(x_2)=0$, the function is only required to be smooth. This contrasts with optimal control theory, where the function must also satisfy a differential equation or system of differential equations. With the exception of low-dimensional special cases, most variational optimisation problems must be solved approximately via numerical methods or basis function expansions.

Variational optimisation can be applied to probabilistic inference problems via the variational lower bound developed in the context of the generalised EM algorithm in section \ref{gemalg}. The key idea is to link the log-likelihood function of the data to a functional of the latent variables, whose distribution is sought. The following treatment mimics the one found in chapter 10 of Bishop \cite{Bishop2009}. The framework is Bayesian, that is, all variables are random, including the parameters, which are described by known prior distributions. Since both the latent variables and the parameters are random, they can be lumped into the variable $Z$. The probabilistic specification is the joint distribution $q(\cdot)$.

\subsection*{Notation}
\begin{itemize}
    \item $X=(x_1,\ldots,x_N),~x_i\in{R}^n$ incomplete data, assumed to be IID (independent and identically distributed).
    \item $Z=(z_1,\ldots,z_N)$ set of missing or latent variables including the unknown parameters, the latter being modelled as random with known prior distributions. The partitioning of the variables does not have to specified, only the joint distribution. The latent variables and parameters do not have to have the same dimensions.
    \item $\calL[q]$ the functional of the joint PDF of latent variables and parameters $q(Z)$.
\end{itemize}

We seek the posterior PDF of the latent variables given the incomplete data ${\rm p}(Z|X)$. Noting that the variable $Z$ contains the latent variables and the parameters, we can write as in the preceding section:
\begin{equation}\label{logpkl2}
    \log {\rm p}(X)=\calL[q]+{\rm KL}(q||p)
\end{equation}
where the functional $\calL[\cdot]$ is defined as
\begin{equation}\label{calL2}
    \calL[q]=\int\log\left(\frac{{\rm p}(X,Z)}{q(Z)}\right)q(Z)\,dZ
\end{equation}
and the Kullback-Liebler divergence is given by
\[
{\rm KL}(q||p)=-\int\log\left(\frac{{\rm p}(Z|X)}{q(Z)}\right)q(Z)\,dZ
\]
which is the same as
\[
{\rm KL}(q||p)=\int\log\left(\frac{q(Z)}{{\rm p}(Z|X)}\right)q(Z)\,dZ
\]
Since the KL divergence is non-negative, in view of (\ref{logpkl2}), $\calL[q]$ is a {\em lower bound} on $\log {\rm p}(X)$:
\begin{equation}
\calL[q]\leq\log {\rm p}(X)
\end{equation}
referred to as the {\em variational lower bound} (VLB). Since the data likelihood ${\rm p}(X)$ is sometimes referred to as the ``evidence'', the bound is also called the {\em evidence lower bound} (ELB or ELBO). Maximising the VLB is equivalent to minimising the KL divergence, which attains the value zero when $q(Z)\equiv{\rm p}(Z|X)$ (equality except on a set of zero measure). The maximisation of this lower bound with respect to the PDF $q(Z)$ requires the solution of a variational optimisation problem, which forms the basis for variational inference. Solving the variational optimisation yields the posterior PDF of the latent variables conditioned on the data. In general, the VLB cannot be obtained explicitly except in certain special cases. It is therefore necessary to resort to approximate techniques. One such approximation is the mean field approximation, covered next.

\subsection{Mean Field Approximation}
Mean field theory (MFT) is an approximate variational technique from physics that has been widely applied in statistical mechanics and field theory \cite{Parisi}. It is an iterative technique that decomposes a multi-dimensional variational optimisation into smaller subproblems by imposing a factorisation on the joint density function $q(Z)$. When applied to variational inference, this type of factorisation is known as the product density transform \cite{Ormerod}. Thus the joint density of the $N$ latent variables (including any parameters, represented by their prior distributions), is assumed to factorise into $M$ factors according to:
\[
q(Z)=\prod_{i=1}^M q_i(Z_i)
\]
which implies a partitioning of the latent variables and parameters as
\[
Z=\bigcup_{i=1}^M Z_i=\{z_1,\ldots,z_N\}
\]
The product density transform reduces the variational optimisation problem to:
\[
\max_{q(\cdot)}\calL[q]\rightarrow
\max_{q_1(\cdot)}\cdots\max_{q_M(\cdot)}\calL[q_1,\ldots,q_M]
\]
This form is somewhat misleading since it implies that the variational optimisation problem is to be solved sequentially on subproblems for the variables $\{Z_1,\ldots,Z_M\}.$ In fact, the mean field approach is iterative: the subproblems are cycled through until convergence. Note also that even if the variational subproblems are solved exactly, the original variational optimisation will not be unless the true joint PDF $q(Z)$ actually factorises in the assumed way.

To illustrate the method, consider the variational subproblem based on variables $Z_j$, which treats all other latent variables $\{Z_i\}$ for $i\neq j$ as constants. We have:
\begin{eqnarray}
\calL[q]&=&\int q_j(Z_j)\,\ln\ptilde(X,Z_j)\,dZ_j
-\int q_j(Z_j)\,\ln q_j(Z_j)\,dZ_j\nonumber\\
&&+~\sum_{i\neq j}\int \ln q_i(Z_i)\prod_{i\neq j}q_i(Z_i)\,dZ_j
\label{Lqj}
\end{eqnarray}
in which the last term does not depend on $q_j(Z_j)$ and we have defined an auxiliary term as:
\begin{equation}\label{ptilde}
\ln\ptilde(X,Z_j)={\rm E}_{\prod_{i\neq j}Z_i}\left[\ln{\rm p}(X,Z)
\right]=``{\rm E}_{i\neq j}\left[\ln{\rm p}(X,Z)
\right]"
\end{equation}
For notational conciseness the latter term is simply written as ``const'' in the sense that it is constant with respect to the argument of the current maximisation  $q_j(Z_j)$. In view of equation (\ref{Lqj}), we can equally write:
\begin{eqnarray}
\calL[q]&=&\int \ln\left(\frac{\ptilde(X,Z_j)}{q_j(Z_j)}\right)
q_j(Z_j)\,\ln\,dZ_j + {\rm const}\nonumber\\
&=&{\rm E}_{Z_j} \ln\left(\frac{\ptilde(X,Z_j)}{q_j(Z_j)}\right) + {\rm const} \nonumber\\
&=& -{\rm KL}(q_j(Z_j)||\ptilde(X,Z_j))+ {\rm const} \label{Lqj2}
\end{eqnarray}

It follows that the first stage of the MFT approximation for variational inference can be expressed as:
\[
\max_{q_j(\cdot)}\calL[q_j,\{q_{i\neq j}\}]=
-{\rm KL}(q_j(Z_j)||\ptilde(X,Z_j))+ {\rm const~w.r.t.}~Z_j
\]
We know that the maximum occurs when the KL divergence attains its minimum value of zero, which is when
\[
q_j^*(Z_j)=\ptilde(X,Z_j))
\]
In view of equation (\ref{ptilde}) we therefore have
\begin{equation}\label{qstar}
q_j^*(Z_j)=\exp\{{\rm E}_{i\neq j}\left[\ln{\rm p}(X,Z)\right]\}
\end{equation}
Since we require all component densities of $q(Z)$ to be PDFs, we must also ensure that (\ref{qstar}) is normalised, leading to
\[
q_j^*(Z_j)=\frac{\exp\{{\rm E}_{i\neq j}\left[\ln{\rm p}(X,Z)\right]\}}
{\int \exp\{{\rm E}_{i\neq j}\left[\ln{\rm p}(X,Z)\right]\}\,dZ_j}
\]

Is is worth noting that the solution for $q_j^*(Z_j)$ is not explicit since it depends on the other variables $Z_{i\neq j}$. The overall algorithm takes the recursive form shown below.
\begin{itemize}
\item[] Initialise the prior densities $q_2^*(Z_2),\ldots,q_M^*(Z_M)$.
\item[] while True do
\begin{itemize}
\item[] for m=1:M
\[
q_m^*(Z_m):=\frac{\exp\{{\rm E}_{i\neq m}\left[\ln{\rm p}(X,Z)\right]\}}
{\int \exp\{{\rm E}_{i\neq m}\left[\ln{\rm p}(X,Z)\right]\}\,dZ_m}
\]
\item[] end for
\end{itemize}
\item[] until convergence criterion on $q(Z)=\prod_{m=1}^M q_m^*(Z_m)$ satisfied
\end{itemize}

In the non-Bayesian version of the MFT approach, we start by specifying a parametric form for the joint density ${\rm p}(Z,X,\Theta)$ where $\Theta$ is the set of unknown parameters and $Z$ are the latent variables. This is obtained by assuming a factored form for the density on partitions of the latent variables. Note that we do not have to assume that the number of factors is $M=N$. The factors are updated in turn according to (\ref{qstar}) and the procedure is iterated until the corrections to the joint PDF become insignificant, or a closed form solution results. The MFT process is guaranteed to converge since the variational lower bound is convex with respect to each of the factors $q_i(Z_i)$ \cite{Boyd}. On the other hand, as in the case of the EM algorithm, the expectations must be explicitly calculable, which restricts the applicability of the approach to classes of log-integrable densities (typically the exponential family).

A more general approach to variational inference is to use a numerical approximation to the expectation over the latent variables in the VLB (\ref{calL2}) so that $\calL[\cdot]$ is optimisable using a gradient-based or other numerical optimisation algorithm.

\section{Variational Autoencoder}\label{aevb}
Autoencoders are unsupervised learning  models that implement a pair of transformations $H(\cdot)$ from input $X$ to latent variable $Z$ and $G(\cdot)$ from latent variable to output $\Xhat$ of the form
\[
Z=H(X),~\Xhat=G(Z)=G(H(X))
\]
such that $\Xhat\approx X$ in some sense (such as sum-of-squares or reconstruction error). When the dimensionality of the latent variable space satisfies ${\rm dim}(Z)\ll{\rm dim}(X)$, the mapping $H$ can be thought of as an encoding to a reduced dimension space $Z$ and $G$ can correspondingly be seen as a decoding. The latter property leads to the terminology of ``autoencoder'' since if $\Xhat=X$, the mappings $G$ and $H$ are ``inverses,'' or at least pseudo-inverses since they map between spaces of unequal dimension. Although any functional representation can be adopted for the encoder $H$ and decoder $G$, artificial neural networks based on multilayer perceptrons (MLPs) have be found to be efficient representations for this task. The problem is then to estimate the parameters of the networks $G$ and $H$, which is typically accomplished via backpropagation or some other numerical optimisation method. Typically, the cost or loss function includes a regularisation term (such as L$_1$ or L$_0$) to ensure an efficient or sparse representation with good generalisation properties: the lower dimensional latent space represents the ``signal'' subspace learnt from the noisy data $X$.

The idea of applying neural networks to this problem dates back at least to 1986 in the work of LeCun \cite{Lecun87}. Autoencoders have found application in dimensionality reduction (for image compression or feature learning), representation learning, de-noising and as generative models. In the latter case $H$ is discarded after the network parameters have been learnt, whereupon $G$ can be used to generate or simulate samples from the data distribution on presentation of a random sample from the latent variable space $Z$.

More recently, two very interesting approaches for designing autoencoders have appeared. The first of these by Kingma and Welling \cite{Kingma2014}, which we cover in detail in this section, is an implementation of variational Bayes called AEVB where the VLB is decomposed as a reconstruction error term and a KL-based regularisation term. The second of these is the generative adversarial network (GAN) framework of Goodfellow et al. \cite{Goodfellow2014}. This is not so much an autoencoder as a generator-discriminator pair. The training framework is game theoretic: the generator implements $\Xhat=G(Z)$ and the discriminator $D$ maps $G(Z)$ to 0 or 1 depending on whether it distinguishes $\Xhat$ as ``real'' or ``fake'' data. Both the AEVB and GAN frameworks have been applied to images via deep convolutional neural networks (CNNs), which form their own feature maps automatically from the data.

\subsection{Reparametrisation Trick}\label{trick}
Before presenting the AEVB algorithm, we deal with a preliminary result that is required in the derivation under the title of the ``reparametrisation trick.'' The result concerns the approximation of an expectation of a function of a random variable under an invertible transformation. Suppose $X$ and $Y$ are two continuous, scalar random variables, related by an invertible transformation $Y=h(Z)$, with inverse transformation $Z=h^{-1}(Y)\keiko g(Y)$. The probability measure is invariant under this transformation in the sense that:
\[
dP_Y(y)=dP_Z(z)
\]
where P denotes the cumulative distribution function ${\rm P}_X(x)={\rm Pr}(X<x)$. In terms of probability density functions, which are assumed to exist, we have
\[
{\rm p}_Y(y)\,dy={\rm p}_Z(z)\,dz
\]
which implies
\[
{\rm p}_Y(y)=\left.\left(\left|\frac{dy}{dz}\right|^{-1}{\rm p}_Z(z)\right)\right|_{z=g(y)}
\]
The expectation of a function of $Z$ can be expressed as:
\begin{equation}\label{reparam1}
{\rm E}(f(Z))=\int f(z)\,{\rm p}_Z(z)\,dz
=\int f(g(y))\,{\rm p}_Y(y)\,dy
\end{equation}
where the integrals are over the real line.
These results also generalise to the vector case when $h:\Reals^n\rightarrow\Reals^n$ is an invertible function with inverse $g(\cdot)$:
\[
{\rm E}(f(Z))=\int\!\!\cdots\!\int f(z)\,{\rm p}_Z(z)\,dz_1\cdots dz_n
=\int\!\!\cdots\!\int f(g(y))\,{\rm p}_Y(y)\,dy_1\cdots dy_n
\]
where the probability measures are related by the Jacobian (i.e. the determinant of the Jacobian matrix) of the transformation:
\begin{equation}\label{chgvarvec}
{\rm p}_Y(y)=\left.\left(\left|\frac{\partial(y_1,\ldots,y_n)}{\partial(z_1,\ldots,z_n)}\right|^{-1}{\rm p}_Z(z)\right)\right|_{z=g(y)}
\end{equation}
The result also generalises to the non-injective case, where multiple $Z$ map to the same $Y$. In this case, all possible solutions to the equation $h(Z)=Y$ must be accounted for by summing over terms on the right side of (\ref{chgvarvec}) (see \cite{Papoulis1991}).

In particular, if $Y_i\sim {\rm p}_Y$, $i=1,\ldots,L$, are IID samples from the distribution of $Y$, a {\em Monte Carlo estimate} of the expectation ${\rm E}(f(Z))$ can be written as (\cite{Bishop2009}, p. 524):
\begin{equation}\label{MCest}
\hat{f}=\frac{1}{L}\sum_{l=1}^L f(g(Y_i))
\end{equation}

\subsection{Autoencoded Variational Bayes (AEVB)}
Despite the high impact of Kingma and Welling's 2014 paper, which has been cited thousands of times (at the time of writing of this tutorial), and the presence of some very good treatments like \cite{Fraccaro2018}, certain aspects of the original paper deserve more attention. In particular, the significance of the key equations in \cite{Kingma2014} can be enhanced by comparing them with the conventional variational Bayes theory in section \ref{VB}. Retaining the notation from the previous section, we turn our attention to this task.

Bearing in mind the definition (\ref{calL2}), for any admissible PDF $q(Z)$, there holds
\begin{eqnarray}
\calL[q]+{\rm KL}(q(Z)||{\rm p}(Z|X))&=&{\rm E}_Z \ln \frac{{\rm p}(X,Z)}{q(Z)}-{\rm E}_Z \ln \frac{{\rm p}(Z|X)}{q(Z)}\label{VLBconv}\\
&=&{\rm E}_Z \ln \frac{{\rm p}(X,Z)}{{\rm p}(Z|X)}\nonumber\\
&=&{\rm E}_Z \ln \frac{{\rm p}(X,Z){\rm p}(X)}{{\rm p}(X,Z)}\nonumber\\
&=&{\rm E}_Z \ln {\rm p}(X)=\ln {\rm p}(X)\nonumber
\end{eqnarray}
which, as we saw previously, gives rise to the variational lower bound on the evidence: $\ln {\rm p}(X)\geq \calL[q]$.

Kingma and Welling employ the decomposition
\[
\ln\frac{{\rm p}(X,Z)}{q(Z)}=\ln\frac{{\rm p}(X|Z){\rm p}(Z)}{q(Z)}
=\ln{\rm p}(X|Z)+\ln\frac{{\rm p}(Z)}{q(Z)}
\]
to re-express the VLB in their equation (3) in the following manner:
\begin{eqnarray}
\calL[q]&=&{\rm E}_Z \ln{\rm p}(X|Z)+{\rm E}_Z\ln\frac{{\rm p}(Z)}{q(Z)}\nonumber\\
&=&{\rm E}_Z \ln{\rm p}(X|Z)-{\rm KL}(q(Z)||{\rm p}(Z))\label{VLBalt}
\end{eqnarray}
We draw attention to the fact that ${\rm p}(Z)$ is the {\em prior} PDF of the latent variables, and not the {\em posterior} as in (\ref{VLBconv}).

For the conventional VLB, we have:
\[
\calL[q]={\rm E}_Z\ln\frac{{\rm p}(X,Z)}{q(Z)}={\rm E}_Z
\left[\ln{\rm p}(X,Z)-\ln q(Z)\right]
\]
Now make use of the ``reparametrization trick'' (\ref{reparam1}), suitably generalised to the vector case with $f(Z)=\ln{\rm p}(X,Z)-\ln q(Z)$ where $q(Z)=q_{\Phi}(Z|X)$ and $Z=g(Y)$, and apply Monte Carlo sampling based on samples $Y_i$ as in (\ref{MCest}) to obtain:
\[
{\rm E} f(Z)\approx\frac{1}{L}\sum_{l=1}^L f(Y_l)
\]
\[
{\rm E} f(Z)\approx\frac{1}{L}\sum_{l=1}^L \ln{\rm p}_{\Theta}(X,Z_l)-\ln q_{\Phi}(Z_l|X)
\]
which justifies formula (6) in Kingma \& Welling:
\begin{equation}\label{calLA}
\calL^A(\Theta,\Phi;X_i)=\frac{1}{L}\sum_{l=1}^L \ln{\rm p}_{\Theta}(X_i,Z_{il})-\ln q_{\Phi}(Z_{il}|X_i)
\end{equation}
in which $Z_{il}=g_{\Phi}(Y_l; X_i)$.

In contrast to this, if we apply the alternative form of the VLB (\ref{VLBalt}), resampling the term ${\rm E}_Z \ln{\rm p}(X|Z)$ with $f(Z)=\ln{\rm p}_{\Theta}(X|Z)$ where $q(Z)=q_{\Phi}(Z|X)$ and $Z=g(Y)$, then we obtain
\begin{equation}\label{calLB}
\calL^B(\Theta,\Phi;X_i)=\frac{1}{L}\sum_{l=1}^L \ln{\rm p}_{\Theta}(X_i|Z_{il})-{\rm KL}(q_{\Phi}(Z|X_i)||{\rm p}_{\Theta}(Z))
\end{equation}
For the Variational Autoencoder (or Variational Auto-Encoder---VAE), Kingma \& Welling assume that the prior ${\rm p}_{\Theta}(Z)$ and the posterior $q_{\Phi}(Z|X_i)$ of the latent variables are both Gaussian. They then apply the transformation $Z_{il}=g_{\Phi}(Y_l;X_i)=\mu_i+\sigma_i\odot Y_l$ to sample the posterior $q_{\Phi}(Z|X_i)$. In this particular case, the KL divergence is easily expressible explicitly in terms of the parameters of the Gaussian PDFs (see equation (10) in (\cite{Kingma2014}). As a consequence, the VLB $\calL^B(\Theta,\Phi;X_i)$ is optimisable by stochastic gradient descent (SGD) on minibatches of training data $\{X_1,\ldots,X_M\}$, $L<M<N$.

Kingma \& Welling mention that in equation (\ref{calLB}), the KL divergence ensures regularisation while the term $\ln{\rm p}_{\Theta}(X_i|Z_{il})$ corresponds to the reconstruction error. Said another way, minimising the KL divergence between the approximate posterior PDF of the latent variables $q_{\Phi}(Z|X_i)$ with respect to the prior PDF ${\rm p}_{\Theta}(Z)$ corresponds to an {\em encoding} operation: the transformation from the input data $X$ to the latent variables $Z$ such that the two PDFs $q_{\Phi}(Z|X)$ and ${\rm p}_{\Theta}(Z)$ are close in the sense of KL divergence. Conversely, minimising the reconstruction error $\ln{\rm p}_{\Theta}(X_i|Z_{il})$ over the set of data samples $\{X_i\}$ is equivalent to producing an estimate $\hat{X}=H(Z,\Theta)$ from the latent variables $Z$, which corresponds to a {\em decoding} operation. Minimising the cost function (\ref{calLB}) therefore amounts to implementing an autoencoder with encoder parameters $\Phi$ and decoder parameters $\Theta$. This justifies the terminology ``autoencoded variational Bayes'' (AEVB). In the 3rd part of their paper, Kingma \& Welling develop a variational autoencoder whose transformations are realised by MLP neural networks. The parameters $\{\Phi, \Theta\}$ are estimated jointly via the AEVB algorithm, which uses SGD on minibatches of data.

\subsection{A Note on Dimensions in the AEVB Algorithm}
The dimensions of the variables throughout  Kingma \& Welling's paper are mostly not specified. While this does not invalidate the arguments, and is in fact commonplace throughout recent AI literature, it is a useful exercise to check the compatibility of the dimensions where this is not stated. A case in point is the reparametrisation trick (section 2.4 of \cite{Kingma2014}), where there is a significant number of variables and it is not obvious how these are related. Quoting K\&W:
``Given the deterministic mapping $z=g_{\phi}(\epsilon,x)$ we know that $q_{\phi}(z|x)\prod_idz_i={\rm p}(\epsilon)\prod_i d\epsilon_i$. Therefore $\int q_{\phi}(z|x) f(z)\, dz=\int {\rm p}(\epsilon) f(g_{\phi}(\epsilon,x))\,d\epsilon$''

While we can see, by substituting $q_{\phi}(z|x)={\rm p}_Z(z)$, $g_{\phi}(\epsilon,x)=g(\epsilon)$ and $\epsilon=y$ in (\ref{reparam1}), that this is an elementary change of (vector) variables, the authors have not actually stated the dimensions of the variables. This raises some ambiguity later in the paper concerning exactly what type of transformation the authors intend. For instance on page 14 of \cite{Kingma2014} concerning the AEVB algorithm, we find
\[
\tilde{z}\sim q_{\phi}(z|x),~\mbox{where}~\tilde{z}=\mu_z+\sigma_z\odot\epsilon,~\epsilon\sim{\cal{N}}(0,I)
\]
Although the authors have not specified the dimensions of any of these variables, the only way to ensure consistency of the equations is to have ${\rm dim}(\mu_z)={\rm dim}(\sigma_z\odot\epsilon)=n_z$ where ${\rm dim}(z)=n_z$. This in turn requires that both $\sigma_z$ and $\epsilon$ are $n_z$-dimensional vectors. The notation for the multivariate Gaussian PDF in their equation (23) is therefore not consistent since $\sigma_z I$ where $I$ is a $n_z\times n_z$ identity matrix, is not a square matrix (which is required to define the covariance of $z$). In fact we need
\[
q_{\phi}(z|x)={\cal{N}}(z;\mu_z,{\rm diag}(\sigma_z\odot\sigma_z))
\]
where diag($v$) is a square matrix of size $n_v\times n_v$, where $n_v={\rm dim}(v)$, whose main diagonal is the entries of the vector $v$ and with zeros elsewhere.
We note that the conditions in section \ref{trick} required for the vector case of equation (\ref{reparam1}) are satisfied. It is also worth noting that the transformation $g_{\phi}(\epsilon,x)$ in section 2.3 of K\&W must be invertible.

\subsection*{Acknowledgements}
The author is grateful to Chafik B. for helpful discussions.

\bibliographystyle{unsrt}
\bibliography{em_vae_refs}

\newpage
\clearpage
\section{Plots \& Tables}

\begin{table}
\center
\begin{tabular}{|c|c|}
\hline
$[\pi_1,\ldots,\pi_k]$ & [0.357, 0.643]\\
\hline
$\mu_1,\ldots,\mu_k$ & 
$\left[
\begin{array}{c}
6.00 \\ 3.02
\end{array}
\right]$
$\left[
\begin{array}{c}
1.89 \\ 1.39
\end{array}
\right]$
\\
\hline
\end{tabular}
\caption{EM results for $\Khat=2$ components. Stop at 20 passes.}
\label{Tab1}
\end{table}

\clearpage

\begin{table}
\center
\begin{tabular}{|c|c|}
\hline
$[\pi_1,\ldots,\pi_k]$ & [0.360, 0.246, 0.394]\\
\hline
$\mu_1,\ldots,\mu_k$ & 
$\left[
\begin{array}{c}
6.00 \\ 3.01
\end{array}
\right]$
$\left[
\begin{array}{c}
-0.01 \\ 2.01
\end{array}
\right]$
$\left[
\begin{array}{c}
3.04 \\ 1.01
\end{array}
\right]$
\\
\hline
$P_1,\ldots,P_k$ & 
$\left[
\begin{array}{cc}
0.50 &-0.02 \\
-0.02 & 0.47
\end{array}
\right]$
$\left[
\begin{array}{cc}
0.48 &0.005 \\
0.005 & 0.46
\end{array}
\right]$
$\left[
\begin{array}{cc}
0.51 & -0.005 \\
-0.005 & 0.49
\end{array}
\right]$
\\
\hline
\end{tabular}
\caption{EM results for $\Khat=3$ components.}
\label{Tab2}
\end{table}

\clearpage

\begin{table}
\center
\begin{tabular}{|c|c|}
\hline
$[\pi_1,\ldots,\pi_k]$ & [0.217, 0.346, 0.078, 0.359]\\
\hline
$\mu_1,\ldots,\mu_k$ & 
$\left[
\begin{array}{c}
-0.04 \\ 2.03
\end{array}
\right]$
$\left[
\begin{array}{c}
3.09 \\ 0.99
\end{array}
\right]$
$\left[
\begin{array}{c}
1.87 \\ 1.41
\end{array}
\right]$
$\left[
\begin{array}{c}
6.00 \\ 3.01
\end{array}
\right]$
\\
\hline
\end{tabular}
\caption{EM results for $\Khat=4$ components.}
\label{Tab3}
\end{table}

\clearpage

\begin{table}
\center
\begin{tabular}{|c|c|}
\hline
$[\pi_1,\ldots,\pi_k]$ & [0.069, 0.004, 0.358, 0.343, 0.226]\\
\hline
$\mu_1,\ldots,\mu_k$ & 
$\left[
\begin{array}{c}
2.14 \\ 1.39
\end{array}
\right]$
$\left[
\begin{array}{c}
3.62 \\ 2.01
\end{array}
\right]$
$\left[
\begin{array}{c}
6.01 \\ 3.01
\end{array}
\right]$
$\left[
\begin{array}{c}
3.08 \\ 0.97
\end{array}
\right]$
$\left[
\begin{array}{c}
-0.06 \\ 2.03
\end{array}
\right]$
\\
\hline
\end{tabular}
\caption{EM results for $\Khat=5$ components. Stop at 50 passes.}
\label{Tab4}
\end{table}

\end{document}